\documentclass{ieeeaccess}
\usepackage{cite}
\usepackage{amsmath,amssymb,amsfonts}
\usepackage{algorithmic}
\usepackage{graphicx}
\usepackage{textcomp}
\usepackage{float}
\usepackage{booktabs}
\usepackage{multirow}
\usepackage{placeins}
\def\BibTeX{{\rm B\kern-.05em{\sc i\kern-.025em b}\kern-.08em
    T\kern-.1667em\lower.7ex\hbox{E}\kern-.125emX}}

\newlength{\xfigwd}
\makeatletter
\newenvironment{accessfigure}[1][!htb]{%
  \@float{figure}[#1]%
  \setlength{\xfigwd}{\columnwidth}%
  \centering
}{%
  \end@float
}
\makeatother
\begin{document}
\history{Date of publication xxxx 00, 0000, date of current version xxxx 00, 0000.}
\doi{10.1109/ACCESS.2017.DOI}

\title{Modeling Art Evaluations from Comparative Judgments: A Deep Learning Approach to Predicting Aesthetic Preferences}
\author{\uppercase{Manoj Reddy Bethi}\authorrefmark{1},
\uppercase{Sai Rupa Jhade}\authorrefmark{1},
\uppercase{Pravallika Yaganti}\authorrefmark{1},
\uppercase{Monoshiz Mahbub Khan}\authorrefmark{1},
and \uppercase{Zhe Yu}\authorrefmark{1}}
\address[1]{Department of Software Engineering, Rochester Institute of Technology, Rochester, NY 14623 USA (e-mail: mb1607@rit.edu, sj7740@rit.edu, yp9365@rit.edu, mk7989@rit.edu, zxyvse@rit.edu)}


\markboth
{Bethi \headeretal: Modeling Art Evaluations from Comparative Judgments}
{Bethi \headeretal: Modeling Art Evaluations from Comparative Judgments}

\corresp{Corresponding author: Zhe Yu (e-mail: zxyvse@rit.edu).}

\begin{abstract}
Modeling human aesthetic judgments in visual art presents significant challenges due to individual preference variability and the high cost of obtaining labeled data. Previous research has shown that models trained on individual ratings often fail to generalize across users, with cross-rater prediction performance lagging behind within-rater outcomes. As a result, when trying to predict ratings from an individual, a sufficient number of high quality training data labels from that individual are always needed. To reduce cost of acquiring such labels, we propose to apply a comparative learning framework based on pairwise preference assessments rather than direct ratings. This approach leverages the Law of Comparative Judgment, which posits that relative choices exhibit less cognitive burden and greater cognitive consistency than direct scoring. We extract deep convolutional features from painting images using ResNet-50 and develop both a deep neural network regression model and a dual-branch pairwise comparison model. We explore four research questions: (RQ1) How does the proposed deep neural network regression model with CNN features compare to the baseline linear regression model using hand-crafted features? (RQ2) How does pairwise comparative learning compare to regression-based prediction when lacking access to direct rating values? (RQ3) Can we predict individual rater preferences through within-rater and cross-rater analysis? (RQ4) What is the annotation cost trade-off between direct ratings and comparative judgments in terms of human time and effort? Our results show that the deep regression model substantially outperforms the baseline, achieving up to 328\% improvement in $R^2$. The comparative model approaches regression performance despite having no access to direct rating values, validating the practical utility of pairwise comparisons. However, predicting individual preferences remains challenging, with both within-rater and cross-rater performance significantly lower than average rating prediction, likely due to rating noise and insufficient individual training data. Human subject experiments (n=5) reveal that comparative judgments require 60\% less annotation time per item, demonstrating superior annotation efficiency for large-scale preference modeling.
\end{abstract}


\titlepgskip=-15pt

\maketitle

\section{Introduction}
\label{sec:introduction}
\PARstart{T}{he} computational modeling of aesthetic preferences in visual art remains a fundamental challenge at the intersection of computer vision, cognitive psychology, and machine learning. While aesthetic experiences are inherently subjective, empirical research has identified consistent patterns in human responses to visual stimuli \cite{b19,b20}. These findings have motivated the development of machine learning models that predict beauty or preference ratings based on visual features extracted from images \cite{b4,b11}.

Recent work by Sidhu et al. \cite{sidhu2018prediction} demonstrated that both subjective features (e.g., emotionality, meaningfulness) and objective image features (e.g., hue, symmetry, brightness) can predict evaluations of abstract and representational paintings. However, most current methodologies rely on direct rating scales, where participants assign numerical values to individual artworks. While straightforward to implement, this approach presents several limitations: (1) participants interpret rating scales inconsistently, introducing noise and variability \cite{b21,b22}; (2) evaluation fatigue and anchoring effects emerge when assessing multiple items \cite{b23}; and (3) models trained on rating data show poor generalization between individuals, with cross-rater prediction performance significantly worse than within-rater prediction \cite{b4,sidhu2018prediction}.

To address these limitations, we propose to apply comparative learning---a preference modeling technique grounded in pairwise evaluations rather than direct ratings. In this framework, participants only need to indicate their preference between two artworks, a task that is more straightforward and cognitively simpler than assigning numerical scores. This concept builds on Thurstone's Law of Comparative Judgment \cite{b2}, which asserts that individuals exhibit greater consistency in relative comparisons than direct ratings. Empirical evidence supports this claim, demonstrating that comparative evaluations reduce response variability and decision fatigue \cite{b25}. Comparative methods have shown success in various aesthetic modeling tasks, including image interestingness \cite{b17}, music emotion labeling \cite{b5}, and human attractiveness prediction \cite{b26}.

\subsection{Research Objectives and Questions}

This research extends previous findings by proposing a comparative learning approach to predict aesthetic preferences using deep neural networks and convolutional features extracted directly from painting images. We address four primary research questions:

\textbf{RQ1: Deep Neural Network Regression vs. Baseline.} How does a deep neural network regression model trained on CNN features (ResNet-50) compare to the baseline linear regression approach using hand-crafted features from Sidhu et al. \cite{sidhu2018prediction}?

\textbf{RQ2: Comparative Learning vs. Regression.} Can a comparative learning model trained on pairwise preference judgments achieve competitive performance compared to regression models trained on direct ratings, despite having no access to the actual rating scale values?

\textbf{RQ3: Within-Rater vs. Cross-Rater Prediction.} Can we successfully predict individual rater preferences? How does within-rater performance (training and testing on the same individual) compare to cross-rater performance (training on other raters, testing on a held-out rater)?

\textbf{RQ4: Annotation Burden Analysis.} What is the annotation burden trade-off between direct ratings and comparative judgments in terms of human time, effort, and data quality? Which method provides better annotation efficiency for large-scale preference modeling?

\subsection{Contributions}

Our primary contributions are:

\begin{enumerate}
    \item \textbf{Baseline Replication:} We replicate the objective-features-only model from Sidhu et al. \cite{sidhu2018prediction} using their 11 hand-crafted features to establish a rigorous baseline.

    \item \textbf{Deep Neural Network Regression:} We develop an improved regression architecture using CNN features extracted from painting images via ResNet-50, with regularization, batch normalization, and dropout that substantially outperforms the baseline.

    \item \textbf{Comparative Learning Framework:} We introduce a pairwise comparison model based on hinge loss optimization, inspired by Khan et al.~\cite{khan2025efficientstorypointestimation}, that approaches regression performance while requiring only relative preference judgments, without access to direct rating values.

    \item \textbf{Comprehensive Evaluation:} We conduct extensive experiments across four conditions (Abstract/Representational $\times$ Beauty/Liking) with within-rater and cross-rater analysis over 10 runs to ensure statistical robustness.

    \item \textbf{Human Survey Validation:} We conduct a human subject study directly comparing annotation burden between direct ratings and comparative judgments. Our findings demonstrate that comparative judgments require 60\% less annotation time per item (10.71s vs 27.28s on average) while maintaining acceptable inter-rater agreement, validating the practical advantages of comparative learning for large-scale preference data collection.
\end{enumerate}

The remainder of this paper is organized as follows: Section II reviews related work, Section III describes our methodology and experimental design, Section IV presents our computational results organized by research question (RQ1-RQ3), Section V presents human survey validation addressing RQ4, Section VI discusses implications and limitations, and Section VII concludes with future directions.

\section{Related Work}

\subsection{Comparative Judgments in Evaluation}

Comparative judgment has a long history across various domains. In education, comparative assessments of students have proven more reliable than individual evaluations \cite{b1}. Verhavert et al. \cite{b3} noted that ``comparative judgments are considered to be easier and more intuitive, as people generally base their decisions on comparisons, either consciously or unconsciously.'' F\"{u}rnkranz and H\"{u}llermeier \cite{b7} similarly argued that pairwise comparisons are more intuitively appealing to human judges, as comparing alternatives is cognitively less demanding than assigning direct values. This observation aligns with Thurstone's Law of Comparative Judgment \cite{b27}, which established the theoretical foundation for preference modeling through pairwise comparisons.

\subsection{Music and Emotional Preference Modeling}

Jensen et al. \cite{b4} developed a predictive model for music preference by asking users to compare pairwise tracks rather than rating individual songs. Using audio features extracted via MFCCs and modeled with Gaussian Mixture Models, they achieved approximately 74\% accuracy on a dataset of 30 music tracks---substantially better than the 50\% random baseline. Madsen et al. \cite{b5} extended this work to predict emotional expressions in music along arousal and valence dimensions using two-choice comparison tasks, Gaussian Processes with probit likelihood, and active learning strategies. While achieving efficient prediction, they acknowledged that personal differences across participants affected outcomes, highlighting the challenge of individual variability in aesthetic judgments.

\subsection{Image Aesthetics and Quality Assessment}

Sidhu et al. \cite{sidhu2018prediction} conducted a comprehensive study measuring four subjective and eleven objective predictors of beauty and liking ratings for 240 abstract and 240 representational paintings. They found that prediction was much better for subjective predictors than objective predictors, and much better for representational than abstract paintings. Critically, they used linear regression with stepwise selection on hand-crafted features including color properties (hue, saturation, brightness) and structural properties (entropy, edge density, symmetry). Their work serves as the baseline for our research.

Our research extends their findings by: (1) using deep CNN features instead of hand-crafted features, (2) implementing deep neural network architectures instead of linear regression, (3) introducing comparative learning as an alternative to ratings, and (4) analyzing within-rater versus cross-rater prediction systematically.

\subsection{Learning from Comparative Judgments}

Fu et al. \cite{b14} addressed the problem of predicting subjective visual qualities from sparse and noisy pairwise data by developing the Unified Robust Learning to Rank (URLR) model, which identifies and corrects outliers while learning to rank images. When applied to image and video datasets, their method significantly increased prediction accuracy. Parekh et al. \cite{b18} used deep pairwise classification and ranking for predicting media interestingness, demonstrating that comparative approaches can effectively capture subjective preferences.

Khan et al. \cite{khan2025efficientstorypointestimation} propose a comparative learning approach to agile story point estimation. This work formats data on backlog items' texts and their direct ratings into a pairwise format to include two items' texts and a comparative label denoting which item has the higher direct rating. This data is used on a single dense layer based encoder structure with Hinge loss to train on the reformatted data for the task of story point estimation to predict an order of the backlog items.

\section{Methodology}

This section presents our comprehensive framework for modeling art evaluations. We investigate three approaches: (1) a baseline regression model replicating Sidhu et al. \cite{sidhu2018prediction} using hand-crafted features, (2) a deep neural network regression model using CNN features, and (3) a comparative learning model based on pairwise preferences using CNN features.

\subsection{Dataset}

We utilize the aesthetic preference dataset introduced by Sidhu et al. \cite{sidhu2018prediction}, consisting of 477 paintings (239 abstract and 238 representational). Five raters independently provided aesthetic evaluations for each painting on two dimensions: beauty and liking, both rated on a scale from 0 to 10.

\textbf{Abstract vs. Representational:} Abstract paintings do not portray obvious semantic content, while representational paintings depict recognizable scenes (primarily landscapes). This distinction is important because Vessel and Rubin \cite{b18vessel} reported greater agreement across individuals regarding the beauty of representational images relative to abstract images, suggesting that semantic content leads to more shared preferences.

\textbf{Beauty vs. Liking:} Following Sidhu et al. \cite{sidhu2018prediction}, we examine both beauty ratings and liking ratings. This distinction is important because some people may like artworks that they do not experience as beautiful.

\subsection{Feature Extraction}

We employ two distinct feature extraction approaches:

\subsubsection{Baseline: Hand-Crafted Features}

For baseline comparison with Sidhu et al. \cite{sidhu2018prediction}, we use their eleven objective visual features computed via MATLAB:

\textbf{Color Features (6):}
\begin{itemize}
    \item HueSD: Standard deviation of hue values
    \item Saturation: Mean color intensity
    \item SaturationSD: Standard deviation of saturation
    \item Brightness: Mean luminance
    \item BrightnessSD: Standard deviation of brightness
    \item ColourComponent: Principal component of RGB channels
\end{itemize}

\textbf{Structural Features (5):}
\begin{itemize}
    \item Entropy: Pixel intensity unpredictability
    \item StraightEdgeDensity: Density of straight line edges
    \item NonStraightEdgeDensity: Density of curved edges
    \item Vertical\_Symmetry: Vertical mirror symmetry score
    \item Horizontal\_Symmetry: Horizontal mirror symmetry score
\end{itemize}

\subsubsection{Deep CNN Features}

For our deep neural network regression and comparative models, we extract features directly from painting images using ResNet-50 \cite{he2016deep} pretrained on ImageNet:

\begin{equation}
\mathbf{x} = \text{ResNet-50}_{\text{avg-pool}}(I) \in \mathbb{R}^{2048}
\end{equation}

where $I$ is the input painting image. We use global average pooling to obtain a 2048-dimensional feature vector. Images are processed in two ways:
\begin{itemize}
    \item \textbf{Origin:} Original image dimensions preserved
    \item \textbf{Resized:} Standardized to 224$\times$224$\times$3 pixels
\end{itemize}

All experiments reported use resized images for consistency with standard ResNet-50 input dimensions.

\subsection{Baseline Regression}

We replicate the regression approach from Sidhu et al. \cite{sidhu2018prediction} using their 11 hand-crafted features. Given features $\mathbf{x} \in \mathbb{R}^{11}$, the baseline predicts aesthetic ratings using ordinary least squares (OLS) regression:

\begin{equation}
\hat{y}_{\text{baseline}} = \mathbf{w}^T \mathbf{x} + b
\end{equation}

where $\mathbf{w} \in \mathbb{R}^{11}$ are learnable weights and $b$ is the bias term. We use all 11 features without stepwise selection to avoid overfitting on our dataset size.

\subsection{Deep Neural Network Regression}

We develop a deep regression architecture using CNN features $\mathbf{x} \in \mathbb{R}^{2048}$ extracted from ResNet-50. The architecture consists of a multi-layer perceptron with batch normalization, dropout regularization, and L2 weight decay. Let $f(\mathbf{x})$ denote our encoder function:

\begin{equation}
\begin{aligned}
\mathbf{h}_1 &= \text{Dropout}_{0.25}(\text{BN}(\text{ReLU}(\mathbf{W}_1 \mathbf{x} + \mathbf{b}_1))) \\
\mathbf{h}_2 &= \text{Dropout}_{0.25}(\text{BN}(\text{ReLU}(\mathbf{W}_2 \mathbf{h}_1 + \mathbf{b}_2))) \\
\mathbf{h}_3 &= \text{BN}(\text{ReLU}(\mathbf{W}_3 \mathbf{h}_2 + \mathbf{b}_3)) \\
\hat{y} &= \mathbf{W}_4 \mathbf{h}_3 + \mathbf{b}_4 = f(\mathbf{x})
\end{aligned}
\end{equation}

where $\mathbf{W}_1 \in \mathbb{R}^{512 \times 2048}$, $\mathbf{W}_2 \in \mathbb{R}^{256 \times 512}$, $\mathbf{W}_3 \in \mathbb{R}^{128 \times 256}$, and $\mathbf{W}_4 \in \mathbb{R}^{1 \times 128}$ are weight matrices.

\begin{accessfigure}[!tbh]
\includegraphics[width=0.8\columnwidth]{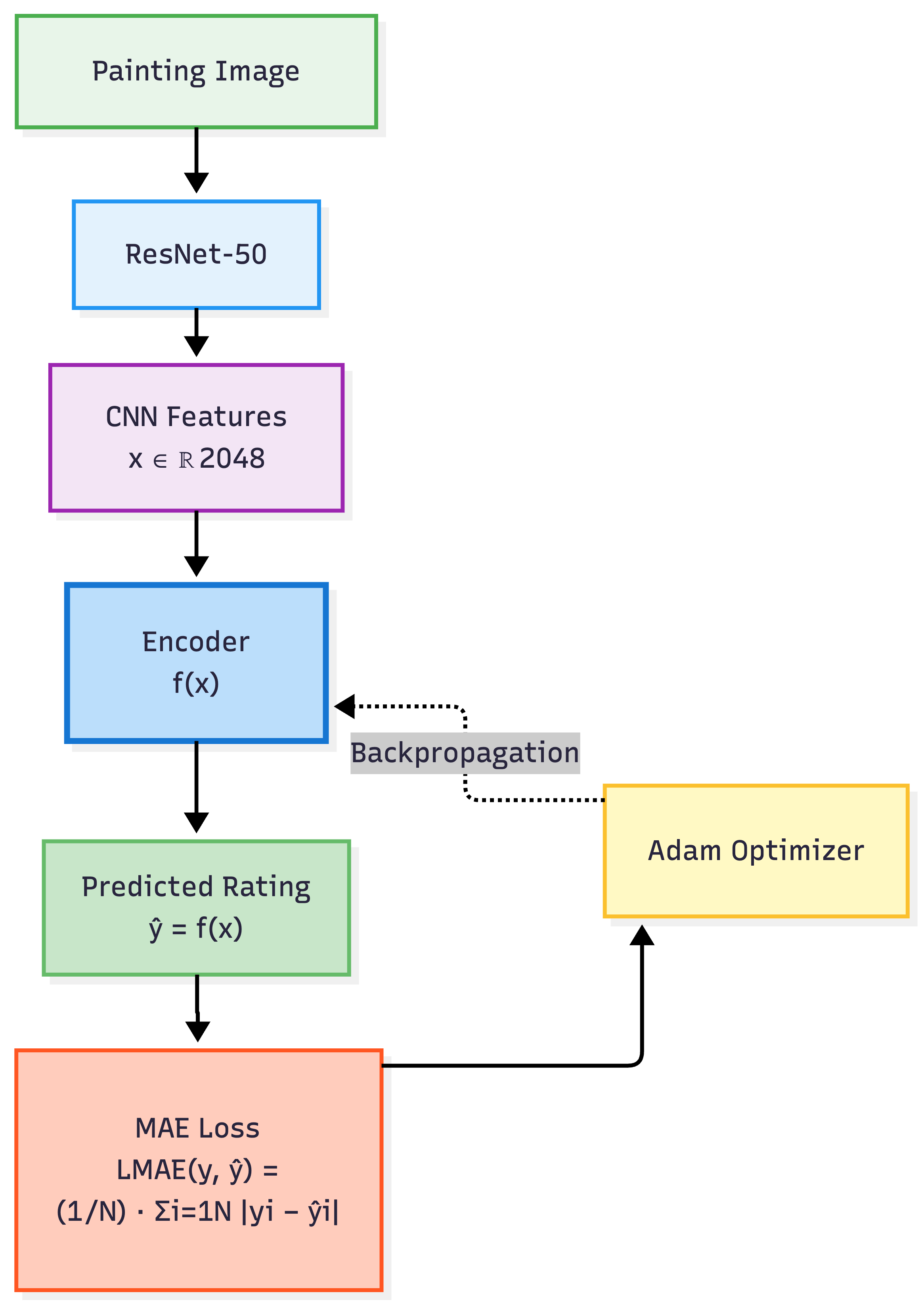}
\caption{Deep neural network regression architecture. Painting images flow from top to bottom: processed through ResNet-50 to extract 2048-dimensional features, fed into encoder $f(\mathbf{x})$ (MLP with BatchNorm and Dropout), and output predicted ratings evaluated with MAE loss.}
\label{fig:regression}
\end{accessfigure}

\textbf{Training:} The model is trained using mean absolute error (MAE) loss with the Adam optimizer:

\begin{equation}
\mathcal{L}_{\text{MAE}} = \frac{1}{N} \sum_{i=1}^{N} |y_i - \hat{y}_i|
\end{equation}

We employ learning rate reduction (factor 0.3, patience 100 epochs) when training loss plateaus, with a minimum learning rate of $10^{-6}$. L2 regularization ($\lambda = 10^{-5}$) is applied to all weight matrices. Training is conducted for 200 epochs with batch size 10.

\subsection{Comparative Learning Model}

Inspired by the comparative learning framework proposed by Khan et al.~\cite{khan2025efficientstorypointestimation}, we use a pairwise comparison approach that learns from relative preferences rather than direct ratings, using the same encoder architecture as our deep regression model.

\subsubsection{Pairwise Preference Generation}

Given two paintings with CNN features $\mathbf{x}_i$ and $\mathbf{x}_j$ and their corresponding ratings $y_i$ and $y_j$, we construct pairwise labels:

\begin{equation}
O_{ij} = \begin{cases}
+1, & \text{if } y_i > y_j \\
-1, & \text{if } y_i < y_j \\
\text{skip}, & \text{if } y_i = y_j
\end{cases}
\end{equation}

For a training set of size $m$, we generate $N$ comparisons per painting, creating up to $N \times m$ pairwise training examples. We vary $N$ from 1 to 10 in our experiments to study the effect of comparison density on model performance.

\subsubsection{Dual-Branch Architecture}

A shared encoder network $f(\mathbf{x})$ (identical to the regression encoder) maps visual features to a latent utility score. For a pair $(\mathbf{x}_i, \mathbf{x}_j)$, we compute:

\begin{equation}
\begin{aligned}
s_i &= f(\mathbf{x}_i) \\
s_j &= f(\mathbf{x}_j) \\
C_{ij} &= s_i - s_j
\end{aligned}
\end{equation}

where $C_{ij}$ represents the predicted preference score.

\begin{accessfigure}[!htb]
\includegraphics[width=0.8\columnwidth]{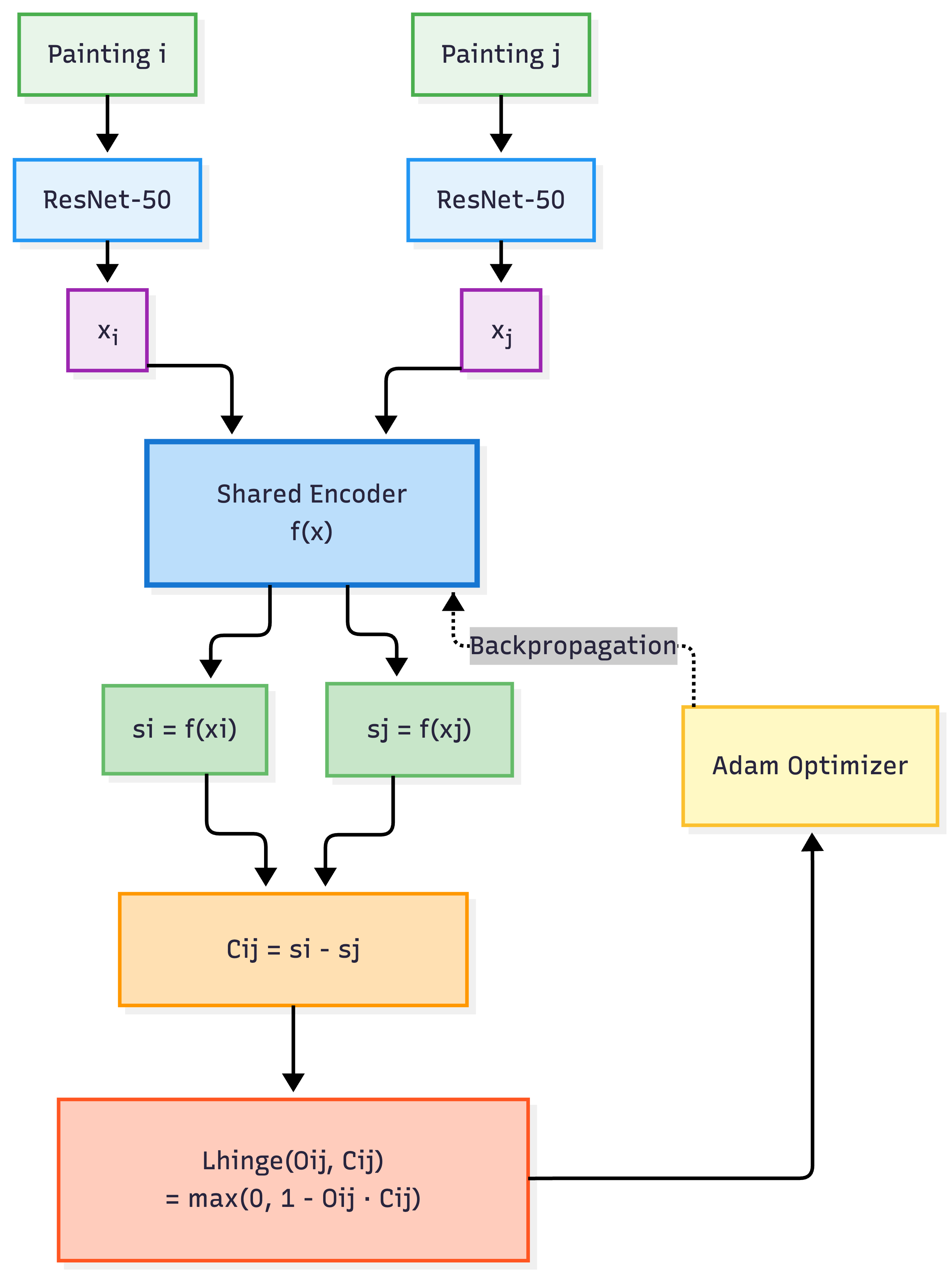}
\caption{Comparative learning framework. Two paintings are processed in parallel from top to bottom: through ResNet-50 for feature extraction, through shared encoder $f(\mathbf{x})$ to produce utility scores $s_i$ and $s_j$, computing difference $C_{ij} = s_i - s_j$, and evaluating against preference label $O_{ij}$ using hinge loss.}
\label{fig:comparative}
\end{accessfigure}

\subsubsection{Hinge Loss Optimization}

We optimize the model using hinge loss, which is well-suited for pairwise ranking:

\begin{equation}
\mathcal{L}_{\text{hinge}}(O_{ij}, C_{ij}) = \max(0, 1 - O_{ij} \cdot C_{ij})
\end{equation}

The total loss across all pairwise comparisons is:

\begin{equation}
\mathcal{L}_{\text{total}} = \frac{1}{|P|} \sum_{(i,j) \in P} \mathcal{L}_{\text{hinge}}(O_{ij}, C_{ij})
\end{equation}

where $P$ is the set of all pairwise comparisons. At test time, we use the learned encoder $f(\mathbf{x})$ to predict direct ratings for unseen paintings. The comparative model is trained for 100 epochs with batch size 10, using the same learning rate schedule and regularization as the regression model.

\subsection{Experimental Settings}

We evaluate all models under three distinct settings:

\textbf{Average Ratings:} The model is trained to predict the average rating across all five raters:
\begin{equation}
\bar{y} = \frac{1}{5} \sum_{r=1}^{5} y_r
\end{equation}

\textbf{Within-Rater:} For each rater $r \in \{1, 2, 3, 4, 5\}$, we train a model exclusively on that rater's data and evaluate on held-out paintings from the same rater. This assesses the model's ability to learn individual aesthetic preferences.

\textbf{Cross-Rater:} For target rater $r$, we train on the average of all other raters and evaluate on rater $r$'s held-out paintings:
\begin{equation}
y_{\text{train}}^{(-r)} = \frac{5 \cdot \bar{y} - y_r}{4}
\end{equation}

This setting evaluates generalization across different individuals.

For all settings, we use a 140/99 random train-test split (approximately 60\%/40\% ratio). We conduct 10 independent runs with different random seeds to ensure statistical robustness, reporting mean performance across runs.

\begin{accessfigure}[!tbh]
\includegraphics[width=0.8\columnwidth]{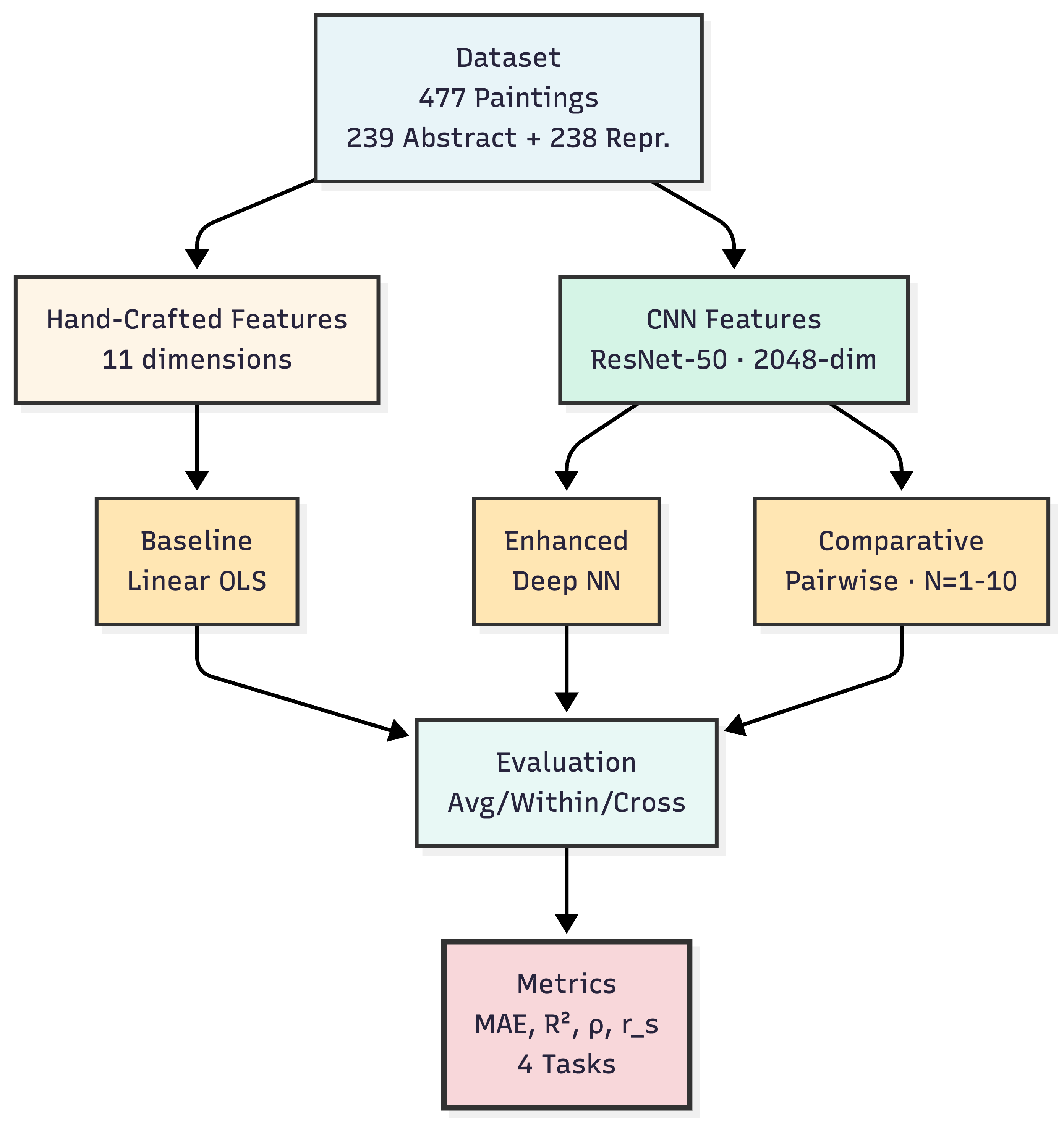}
\caption{Overall methodology framework. The dataset flows from top to bottom through two feature extraction pipelines (hand-crafted and CNN features), three modeling approaches (Baseline OLS, Enhanced Deep NN, Comparative Pairwise), unified evaluation settings (Average/Within-rater/Cross-rater), and final metrics across four aesthetic tasks.}
\label{fig:methodology}
\end{accessfigure}
\FloatBarrier

\subsection{Evaluation Metrics}

Model performance is evaluated using four complementary metrics:

\textbf{Mean Absolute Error (MAE):}
\begin{equation}
\text{MAE} = \frac{1}{n} \sum_{i=1}^{n} |y_i - \hat{y}_i|
\end{equation}

\textbf{Coefficient of Determination ($R^2$):}
\begin{equation}
R^2 = 1 - \frac{\sum_{i=1}^{n} (y_i - \hat{y}_i)^2}{\sum_{i=1}^{n} (y_i - \bar{y})^2}
\end{equation}

\textbf{Pearson Correlation Coefficient ($\rho$):}
\begin{equation}
\rho = \frac{\sum_{i=1}^{n} (y_i - \bar{y})(\hat{y}_i - \bar{\hat{y}})}{\sqrt{\sum_{i=1}^{n} (y_i - \bar{y})^2} \sqrt{\sum_{i=1}^{n} (\hat{y}_i - \bar{\hat{y}})^2}}
\end{equation}

\textbf{Spearman Rank Correlation ($r_s$):}
\begin{equation}
r_s = 1 - \frac{6 \sum_{i=1}^{n} d_i^2}{n(n^2 - 1)}
\end{equation}

where $d_i$ is the difference between ranks of $y_i$ and $\hat{y}_i$. Higher values indicate better prediction performance.

\section{Results}

We organize our results according to our four research questions, presenting comprehensive findings across all four experimental conditions (Abstract Beauty, Abstract Liking, Representational Beauty, Representational Liking). RQ1-RQ3 address computational model performance, while RQ4 addresses annotation burden based on human subject experiments.

\subsection{RQ1: Deep Neural Network Regression vs. Baseline}

Table \ref{tab:baseline_vs_regression} compares the baseline linear regression model using 11 hand-crafted features with our deep neural network regression model using 2048-dimensional CNN features on average ratings.

\begin{table}[h]
\centering
\caption{Baseline vs. Deep Neural Network Regression: Average Ratings}
\label{tab:baseline_vs_regression}
\footnotesize
\begin{tabular}{l|cc|cc|cc}
\toprule
\textbf{Task} & \multicolumn{2}{c|}{\textbf{$R^2$}} & \multicolumn{2}{c|}{\textbf{Pearson $\rho$}} & \multicolumn{2}{c}{\textbf{Spearman $r_s$}} \\
& Base & Deep & Base & Deep & Base & Deep \\
\midrule
Abstract Beauty & 0.090 & \textbf{0.385} & 0.299 & \textbf{0.649} & 0.276 & \textbf{0.594} \\
Abstract Liking & 0.275 & 0.277 & 0.524 & \textbf{0.563} & 0.485 & \textbf{0.502} \\
Repr. Beauty & 0.283 & \textbf{0.344} & 0.532 & \textbf{0.631} & 0.526 & \textbf{0.617} \\
Repr. Liking & 0.335 & \textbf{0.429} & 0.579 & \textbf{0.666} & 0.601 & \textbf{0.658} \\
\bottomrule
\end{tabular}
\end{table}

\textbf{Key Findings:}

The deep neural network regression model substantially outperforms the baseline across nearly all metrics and tasks:

\begin{itemize}
    \item \textbf{Abstract Beauty} shows the largest improvement: $R^2$ increases from 0.090 to 0.385 (328\% relative improvement), Pearson $\rho$ from 0.299 to 0.649 (117\% improvement).

    \item \textbf{Abstract Liking} shows minimal improvement: $R^2$ 0.275 to 0.277, suggesting liking judgments for abstract art are particularly difficult to predict even with rich CNN features.

    \item \textbf{Representational Beauty}: $R^2$ improves from 0.283 to 0.344 (22\% improvement), Pearson $\rho$ from 0.532 to 0.631 (19\% improvement).

    \item \textbf{Representational Liking}: $R^2$ improves from 0.335 to 0.429 (28\% improvement), Pearson $\rho$ from 0.579 to 0.666 (15\% improvement).
\end{itemize}

\textbf{Conclusion for RQ1:} The deep neural network regression model using CNN features significantly outperforms the baseline linear regression using hand-crafted features. The improvement is most pronounced for Abstract Beauty (328\% in $R^2$) and substantial for representational paintings. This demonstrates that deep CNN features capture richer visual representations than hand-crafted features, and non-linear modeling through deep networks better captures the complex relationships between visual features and aesthetic judgments.

\subsection{RQ2: Comparative Learning vs. Deep Regression}

Figures \ref{fig:abstract_beauty_comp} through \ref{fig:representational_liking_comp} compare pairwise comparative model performance against the deep neural network regression baseline as the number of comparison pairs per painting ($N$) increases from 1 to 10.

\begin{accessfigure}[!htb]
\includegraphics[width=1.0\columnwidth]{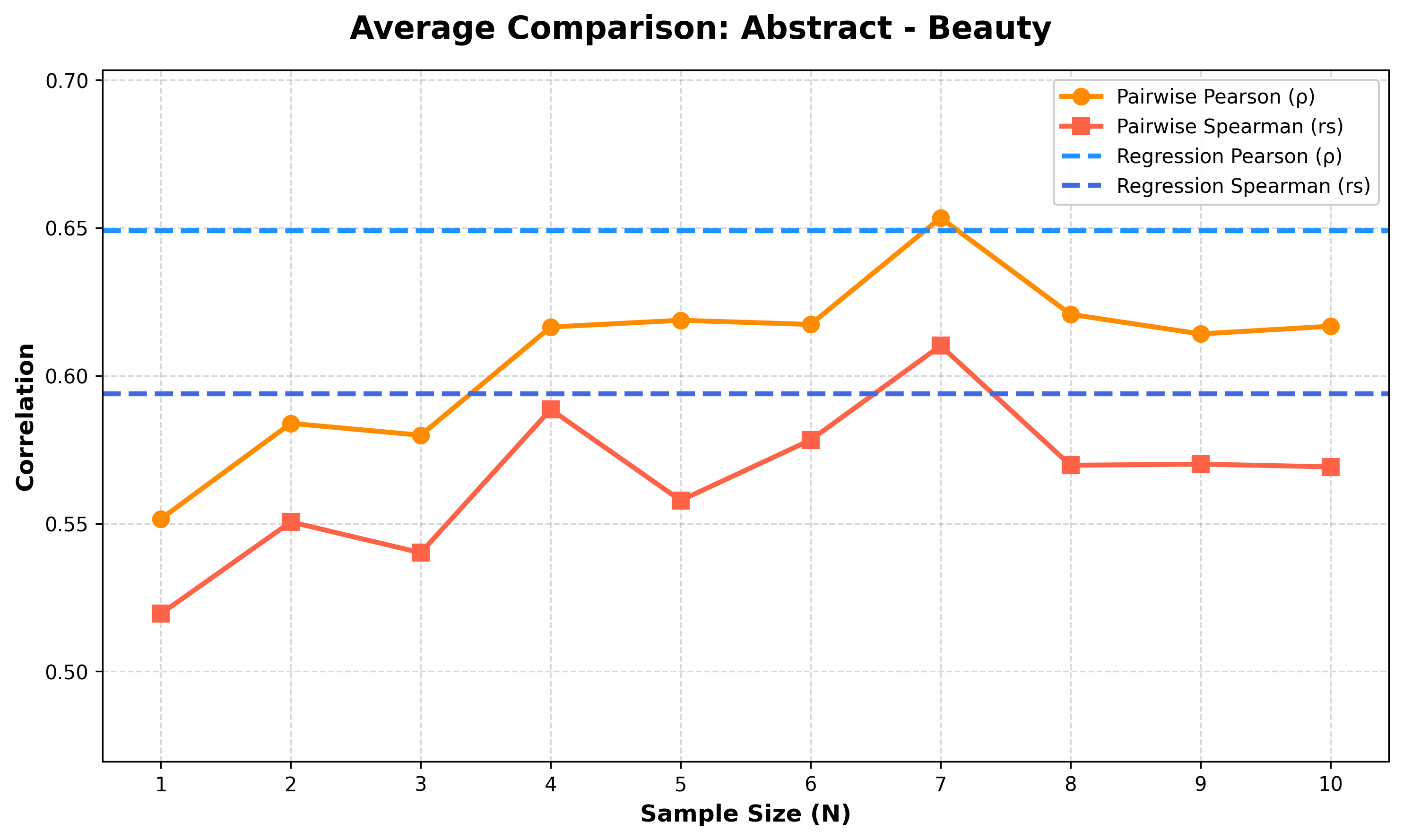}
\caption{Comparative vs. Regression: Abstract Beauty. Pairwise model performance improves with increased sample size $N$ (number of comparison pairs per painting) but remains below regression baseline.}
\label{fig:abstract_beauty_comp}
\end{accessfigure}
\FloatBarrier

\begin{accessfigure}[!htb]
\includegraphics[width=1.0\columnwidth]{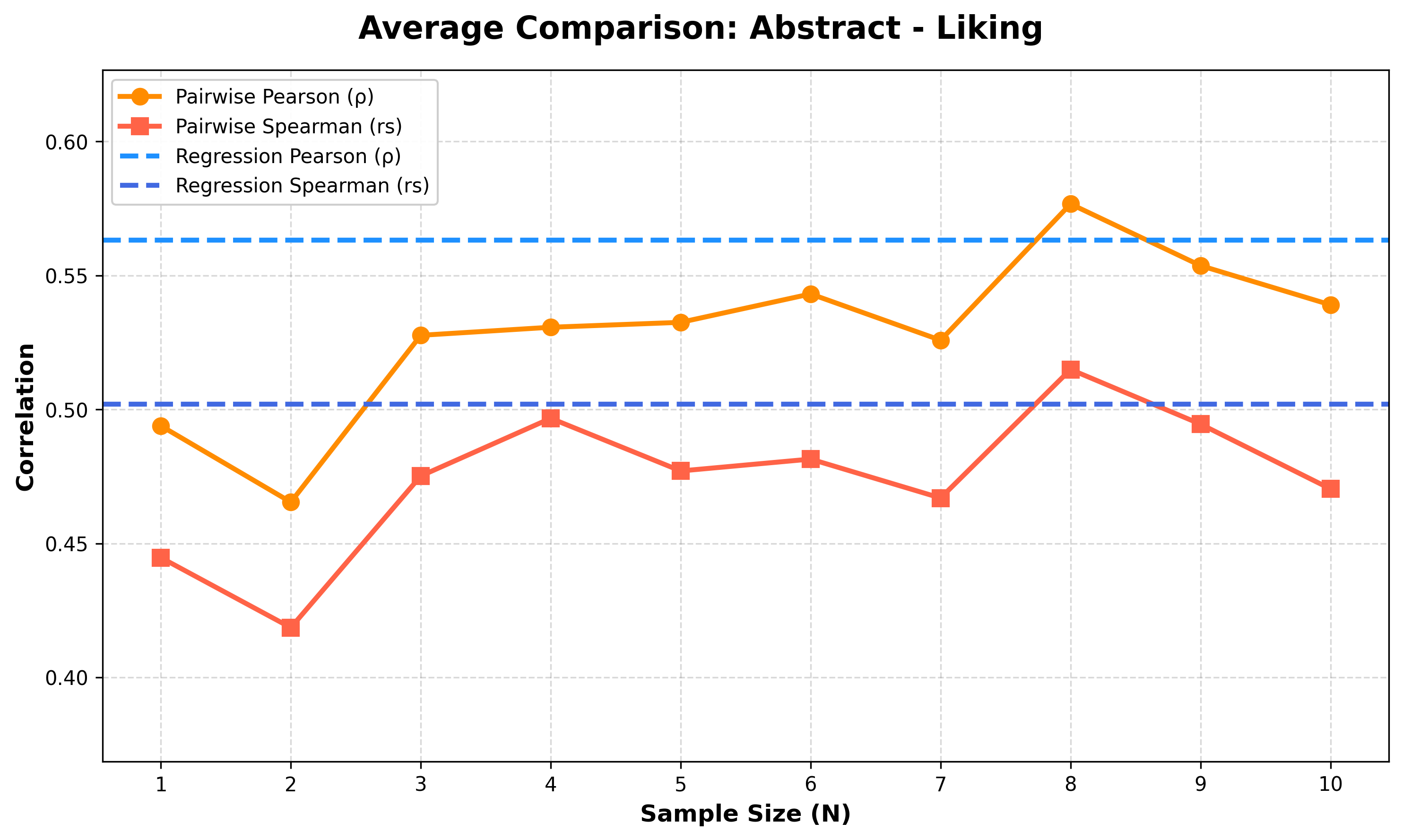}
\caption{Comparative vs. Regression: Abstract Liking. Comparative learning shows fluctuating performance below regression baseline.}
\label{fig:abstract_liking_comp}
\end{accessfigure}
\FloatBarrier

\begin{accessfigure}[!htb]
\includegraphics[width=1.0\columnwidth]{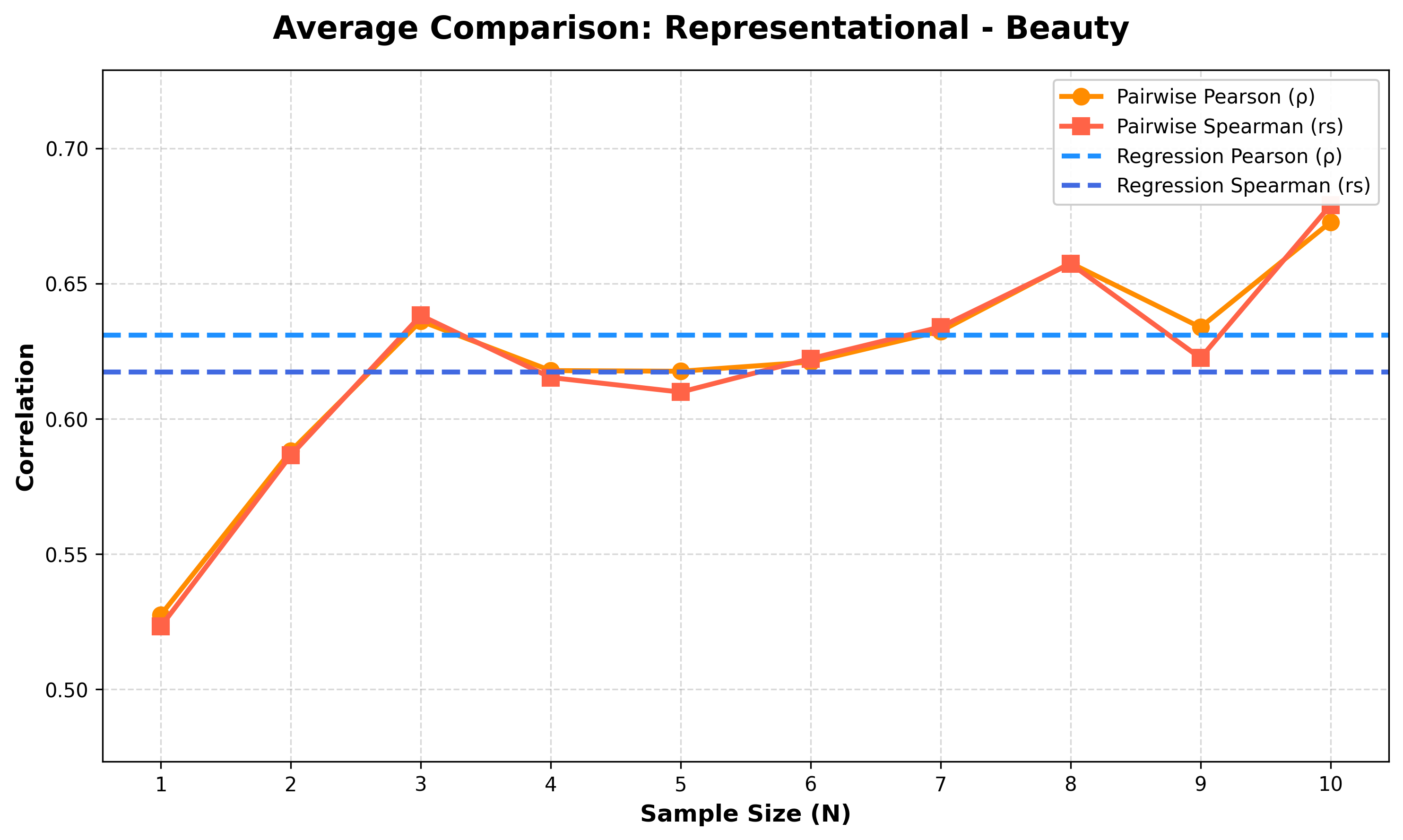}
\caption{Comparative vs. Regression: Representational Beauty. Pairwise model approaches regression performance at higher sample sizes.}
\label{fig:representational_beauty_comp}
\end{accessfigure}
\FloatBarrier

\begin{accessfigure}[!htb]
\includegraphics[width=1.0\columnwidth]{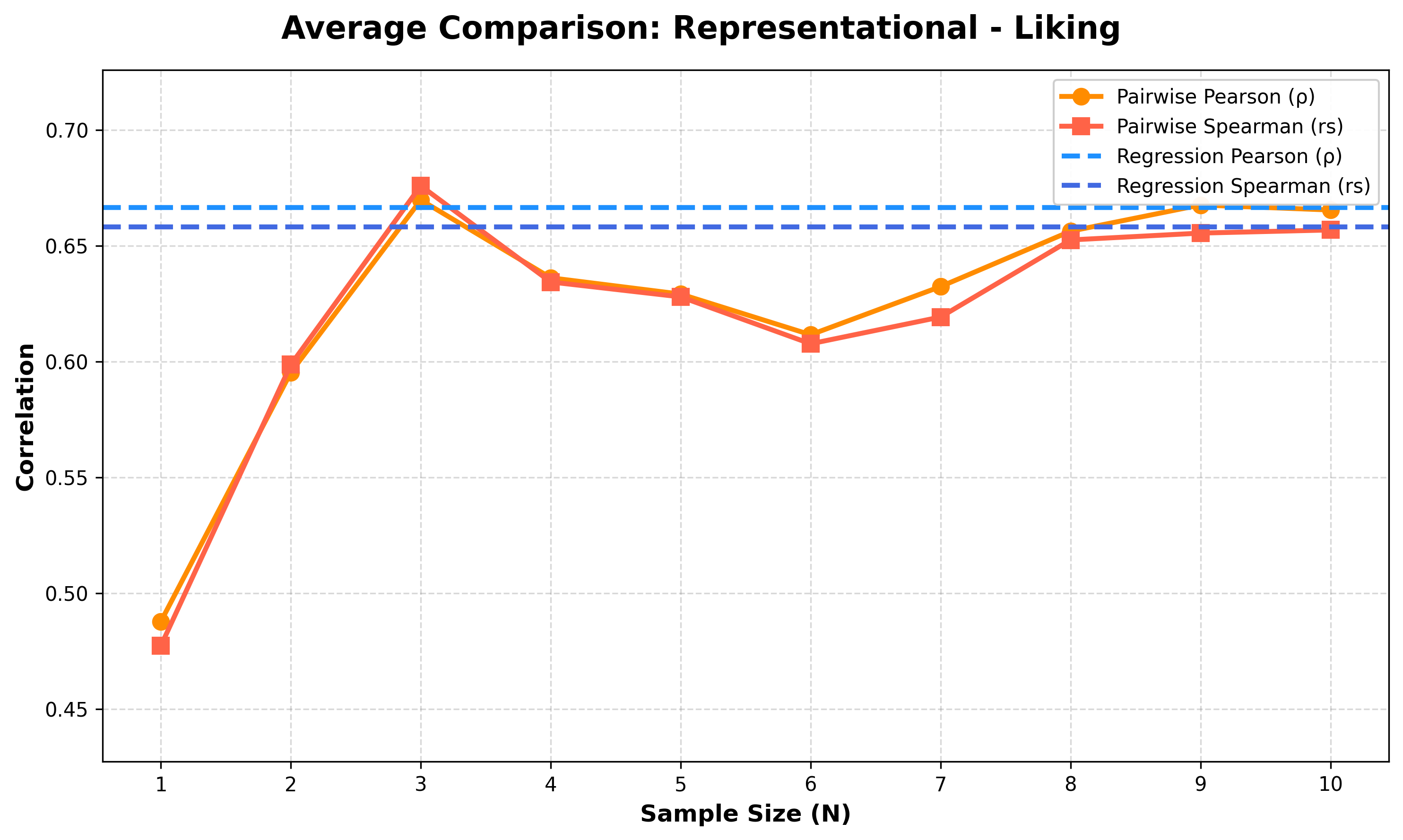}
\caption{Comparative vs. Regression: Representational Liking. Comparative learning gets close to regression performance at N=10.}
\label{fig:representational_liking_comp}
\end{accessfigure}
\FloatBarrier

\textbf{Key Findings:}

\begin{itemize}
    \item \textbf{Performance Gap:} The comparative model consistently performs below the regression model across all tasks, which is expected since it lacks access to direct rating values and learns only from relative preferences.

    \item \textbf{Sample Size Effect:} Performance generally improves as $N$ increases, demonstrating that more pairwise comparisons provide richer training signal. However, gains plateau around $N=7$ to $10$.

    \item \textbf{Representational Paintings:} Comparative models achieve their best performance on representational paintings, approaching within 5-10\% of regression performance at $N=10$.

    \item \textbf{Abstract Paintings:} Abstract paintings remain more challenging, with larger gaps between comparative and regression performance, particularly for Abstract Liking.
\end{itemize}

\textbf{Conclusion for RQ2:} The comparative learning model performs below the regression model, as expected given that it does not have access to direct rating values. However, it successfully approaches regression performance (within 5-10\%) for representational paintings at higher sample sizes ($N \geq 7$), demonstrating that pairwise preferences can capture much of the information present in direct ratings. This validates the practical utility of comparative judgments as an easier and more reliable annotation method, accepting a modest performance trade-off for significantly reduced cognitive burden on raters and faster data collection.

\subsection{RQ3: Within-Rater vs. Cross-Rater Prediction}

Tables \ref{tab:abstract_beauty_rater} through \ref{tab:representational_liking_rater} present within-rater and cross-rater results for all models across all five raters, averaged over 10 runs.

\begin{table}[h!]
\centering
\caption{Within-Rater vs. Cross-Rater: Abstract Beauty}
\label{tab:abstract_beauty_rater}
\small
\begin{tabular}{|l|cc|cc|}
\hline
\textbf{Model} & \multicolumn{2}{c|}{\textbf{Within-Rater}} & \multicolumn{2}{c|}{\textbf{Cross-Rater}} \\
\hline
& $\rho$ & $r_s$ & $\rho$ & $r_s$ \\
\hline
Baseline & 0.265 & 0.267 & 0.227 & 0.216 \\
Deep NN Regression & 0.404 & 0.384 & 0.359 & 0.316 \\
\hline
Comp. N=1 & 0.281 & 0.272 & 0.239 & 0.208 \\
Comp. N=2 & 0.314 & 0.295 & 0.254 & 0.228 \\
Comp. N=3 & 0.337 & 0.318 & 0.261 & 0.233 \\
Comp. N=4 & 0.332 & 0.316 & 0.294 & 0.259 \\
Comp. N=5 & 0.338 & 0.320 & 0.288 & 0.262 \\
Comp. N=6 & 0.330 & 0.314 & 0.277 & 0.248 \\
Comp. N=7 & 0.320 & 0.296 & 0.288 & 0.253 \\
Comp. N=8 & 0.339 & 0.327 & 0.300 & 0.269 \\
Comp. N=9 & 0.316 & 0.304 & 0.275 & 0.245 \\
Comp. N=10 & 0.323 & 0.303 & 0.287 & 0.262 \\
\hline
\end{tabular}
\end{table}

\begin{table}[h!]
\centering
\caption{Within-Rater vs. Cross-Rater: Abstract Liking}
\label{tab:abstract_liking_rater}
\small
\begin{tabular}{|l|cc|cc|}
\hline
\textbf{Model} & \multicolumn{2}{c|}{\textbf{Within-Rater}} & \multicolumn{2}{c|}{\textbf{Cross-Rater}} \\
\hline
& $\rho$ & $r_s$ & $\rho$ & $r_s$ \\
\hline
Baseline & 0.207 & 0.192 & 0.022 & -0.005 \\
Deep NN Regression & 0.271 & 0.258 & 0.203 & 0.176 \\
\hline
Comp. N=1 & 0.201 & 0.204 & 0.104 & 0.095 \\
Comp. N=2 & 0.225 & 0.222 & 0.094 & 0.080 \\
Comp. N=3 & 0.271 & 0.260 & 0.138 & 0.120 \\
Comp. N=4 & 0.247 & 0.239 & 0.132 & 0.110 \\
Comp. N=5 & 0.234 & 0.225 & 0.112 & 0.101 \\
Comp. N=6 & 0.260 & 0.253 & 0.111 & 0.099 \\
Comp. N=7 & 0.213 & 0.208 & 0.129 & 0.115 \\
Comp. N=8 & 0.243 & 0.237 & 0.115 & 0.093 \\
Comp. N=9 & 0.248 & 0.248 & 0.123 & 0.106 \\
Comp. N=10 & 0.241 & 0.238 & 0.098 & 0.087 \\
\hline
\end{tabular}
\end{table}

\begin{table}[h!]
\centering
\caption{Within-Rater vs. Cross-Rater: Representational Beauty}
\label{tab:representational_beauty_rater}
\small
\begin{tabular}{|l|cc|cc|}
\hline
\textbf{Model} & \multicolumn{2}{c|}{\textbf{Within-Rater}} & \multicolumn{2}{c|}{\textbf{Cross-Rater}} \\
\hline
& $\rho$ & $r_s$ & $\rho$ & $r_s$ \\
\hline
Baseline & 0.105 & 0.110 & 0.154 & 0.149 \\
Deep NN Regression & 0.341 & 0.330 & 0.389 & 0.385 \\
\hline
Comp. N=1 & 0.258 & 0.240 & 0.322 & 0.318 \\
Comp. N=2 & 0.290 & 0.275 & 0.333 & 0.330 \\
Comp. N=3 & 0.280 & 0.272 & 0.366 & 0.356 \\
Comp. N=4 & 0.287 & 0.275 & 0.393 & 0.384 \\
Comp. N=5 & 0.284 & 0.268 & 0.350 & 0.338 \\
Comp. N=6 & 0.284 & 0.283 & 0.376 & 0.364 \\
Comp. N=7 & 0.292 & 0.279 & 0.357 & 0.342 \\
Comp. N=8 & 0.289 & 0.278 & 0.390 & 0.380 \\
Comp. N=9 & 0.272 & 0.258 & 0.383 & 0.375 \\
Comp. N=10 & 0.275 & 0.267 & 0.380 & 0.368 \\
\hline
\end{tabular}
\end{table}

\begin{table}[h!]
\centering
\caption{Within-Rater vs. Cross-Rater: Representational Liking}
\label{tab:representational_liking_rater}
\small
\begin{tabular}{|l|cc|cc|}
\hline
\textbf{Model} & \multicolumn{2}{c|}{\textbf{Within-Rater}} & \multicolumn{2}{c|}{\textbf{Cross-Rater}} \\
\hline
& $\rho$ & $r_s$ & $\rho$ & $r_s$ \\
\hline
Baseline & 0.091 & 0.084 & 0.049 & 0.031 \\
Deep NN Regression & 0.262 & 0.255 & 0.241 & 0.242 \\
\hline
Comp. N=1 & 0.182 & 0.179 & 0.162 & 0.166 \\
Comp. N=2 & 0.207 & 0.203 & 0.167 & 0.172 \\
Comp. N=3 & 0.198 & 0.194 & 0.164 & 0.166 \\
Comp. N=4 & 0.237 & 0.231 & 0.162 & 0.164 \\
Comp. N=5 & 0.230 & 0.217 & 0.172 & 0.166 \\
Comp. N=6 & 0.245 & 0.242 & 0.192 & 0.188 \\
Comp. N=7 & 0.236 & 0.234 & 0.162 & 0.165 \\
Comp. N=8 & 0.207 & 0.200 & 0.169 & 0.165 \\
Comp. N=9 & 0.230 & 0.225 & 0.175 & 0.178 \\
Comp. N=10 & 0.234 & 0.224 & 0.172 & 0.173 \\
\hline
\end{tabular}
\end{table}

\textbf{Key Findings:}

\begin{itemize}
    \item \textbf{Overall Low Performance:} Both within-rater and cross-rater performance are significantly lower than average rating prediction (compare $\rho = 0.27$--$0.40$ for individuals vs. $\rho = 0.56$--$0.67$ for average ratings in Table \ref{tab:baseline_vs_regression}).

    \item \textbf{Within-Rater Performance:} Even when training and testing on the same rater, performance remains modest (Pearson $\rho$ ranging from 0.20 to 0.40), indicating high noise or inconsistency in individual ratings.

    \item \textbf{Cross-Rater Challenge:} Cross-rater performance is generally equal to or slightly lower than within-rater performance, confirming that aesthetic preferences do not generalize well across individuals.

    \item \textbf{Exception - Representational Beauty:} Interestingly, for Representational Beauty, cross-rater performance ($\rho = 0.389$) actually exceeds within-rater performance ($\rho = 0.341$) for the deep regression model, suggesting that aggregate preferences may be more stable than individual judgments for this specific dimension.

    \item \textbf{Abstract Liking Hardest:} Abstract Liking shows the poorest performance across all settings, with baseline cross-rater achieving near-zero correlation ($\rho = 0.022$), confirming this is the most idiosyncratic judgment type.
\end{itemize}

\textbf{Conclusion for RQ3:} Predicting individual rater preferences is not feasible with the available data and methodology. Both within-rater and cross-rater performance remain significantly lower than average rating prediction, indicating that:

\begin{enumerate}
    \item \textbf{High Rating Noise:} Individual ratings contain substantial noise or inconsistency, making them difficult to predict even when training on the same rater's data.

    \item \textbf{Insufficient Data:} With only $\sim$140 training examples per rater, there is insufficient data to learn the unique preference patterns of individual raters. Personalized aesthetic models would likely require orders of magnitude more data per individual.

    \item \textbf{Idiosyncratic Preferences:} The low cross-rater performance confirms that aesthetic preferences are highly individual and do not transfer well between raters, particularly for abstract art and liking judgments.

    \item \textbf{Average Ratings More Stable:} The substantially better performance on average ratings (averaging across 5 raters) suggests that aggregation reduces noise and captures more stable, generalizable aesthetic properties.
\end{enumerate}

\section{Human Survey Validation (RQ4)}

To address RQ4 (annotation burden trade-offs) and validate the practical advantages of comparative judgments, we conducted a human survey directly comparing direct rating and comparative judgment methods.

\subsection{Survey Design}

Seven participants completed aesthetic evaluations using both methods. After variance filtering to remove disengaged participants (those giving constant responses), five participants showed adequate response variability and were retained for analysis. Each participant evaluated paintings using:

\textbf{Direct Ratings:} Participants rated 10 paintings (5 abstract, 5 representational) on a 1-10 scale for both liking and beauty.

\textbf{Comparative Judgments:} Participants made 10 pairwise comparisons (5 abstract, 5 representational), choosing which painting was more beautiful or likable.

\subsection{Analysis Methodology}

To ensure fair comparison, we converted direct ratings to pairwise preferences using the same structure as comparative judgments. For each painting pair $(i,j)$:

\begin{equation}
\text{preference}_{ij} =
\begin{cases}
+1 & \text{if } \text{rating}_i > \text{rating}_j \\
-1 & \text{if } \text{rating}_i < \text{rating}_j
\end{cases}
\end{equation}

Painting pairs with equal ratings were excluded from analysis. We measured inter-rater agreement using Cohen's Kappa and Pearson correlation, averaging across all participant pairs. Cohen's Kappa values range from -1 (perfect disagreement) to +1 (perfect agreement), with 0 indicating chance-level agreement. Only participants showing variance in both methods were included to ensure fair comparison.

\subsection{Results}

Table \ref{tab:human_survey_summary} presents a summary comparison of annotation time between direct ratings and comparative judgments across all four conditions.

\begin{table}[h!]
\centering
\caption{Human Survey: Inter-Rater Agreement Comparison (n=5)}
\label{tab:human_survey_summary}
\small
\begin{tabular}{lcc}
\hline
\textbf{Condition} & \textbf{Method} & \textbf{Time (s)} \\
\hline
Abstract Beauty & Direct & 25.32 \\
 & Comparative & \textbf{13.89} \\
Abstract Liking & Direct & 25.32 \\
 & Comparative & \textbf{13.89} \\
Repr. Beauty & Direct & 29.23 \\
 & Comparative & \textbf{7.53} \\
Repr. Liking & Direct & 29.23 \\
 & Comparative & \textbf{7.53} \\
\hline
\end{tabular}
\end{table}

Tables \ref{tab:abstract_beauty_matrix} through \ref{tab:repr_liking_matrix} present detailed per-rater accuracy values for both direct ratings and comparative judgments. Each table shows a matrix where rows represent the method used by each rater, and columns show agreement accuracy with other raters (GT, R1, R3, R4, R5, R6, and the Average across all comparisons). Rater R2 (P2) is excluded due to insufficient response variance.

\begin{table}[!t]
\centering
\caption{Inter-Rater Accuracy Matrix: Abstract Beauty}
\label{tab:abstract_beauty_matrix}
\scriptsize
\setlength{\tabcolsep}{3pt}
\begin{tabular}{|l|l|cccccc|c|}
\hline
\multicolumn{2}{|c|}{} & \textbf{GT} & \textbf{R1} & \textbf{R3} & \textbf{R4} & \textbf{R5} & \textbf{R6} & \textbf{Avg} \\
\hline
\textbf{GT} & Direct & 1.000 & 0.710 & 0.440 & 0.400 & 0.120 & 0.300 & 0.495 \\
 & Comparative & 1.000 & 0.200 & 0.000 & 0.200 & 0.800 & 0.400 & 0.433 \\
\hline
\textbf{R1} & Direct & 0.710 & 1.000 & 0.830 & 0.430 & 0.200 & 0.570 & 0.623 \\
 & Comparative & 0.200 & 1.000 & 0.800 & 1.000 & 0.400 & 0.800 & 0.700 \\
\hline
\textbf{R3} & Direct & 0.440 & 0.830 & 1.000 & 0.780 & 0.710 & 0.890 & 0.775 \\
 & Comparative & 0.000 & 0.800 & 1.000 & 0.800 & 0.200 & 0.600 & 0.567 \\
\hline
\textbf{R4} & Direct & 0.400 & 0.430 & 0.780 & 1.000 & 0.880 & 0.700 & 0.698 \\
 & Comparative & 0.200 & 1.000 & 0.800 & 1.000 & 0.400 & 0.800 & 0.700 \\
\hline
\textbf{R5} & Direct & 0.120 & 0.200 & 0.710 & 0.880 & 1.000 & 0.880 & 0.632 \\
 & Comparative & 0.800 & 0.400 & 0.200 & 0.400 & 1.000 & 0.200 & 0.500 \\
\hline
\textbf{R6} & Direct & 0.300 & 0.570 & 0.890 & 0.700 & 0.880 & 1.000 & 0.723 \\
 & Comparative & 0.400 & 0.800 & 0.600 & 0.800 & 0.200 & 1.000 & 0.633 \\
\hline
\end{tabular}
\end{table}

\begin{table}[!t]
\centering
\caption{Inter-Rater Accuracy Matrix: Abstract Liking}
\label{tab:abstract_liking_matrix}
\scriptsize
\setlength{\tabcolsep}{3pt}
\begin{tabular}{|l|l|cccccc|c|}
\hline
\multicolumn{2}{|c|}{} & \textbf{GT} & \textbf{R1} & \textbf{R3} & \textbf{R4} & \textbf{R5} & \textbf{R6} & \textbf{Avg} \\
\hline
\textbf{GT} & Direct & 1.000 & 0.560 & 0.670 & 0.300 & 0.400 & 0.330 & 0.543 \\
 & Comparative & 1.000 & 0.600 & 0.400 & 0.200 & 0.400 & 0.600 & 0.533 \\
\hline
\textbf{R1} & Direct & 0.560 & 1.000 & 0.620 & 0.440 & 0.560 & 0.380 & 0.593 \\
 & Comparative & 0.600 & 1.000 & 0.400 & 0.200 & 0.400 & 0.600 & 0.533 \\
\hline
\textbf{R3} & Direct & 0.670 & 0.620 & 1.000 & 0.670 & 0.670 & 0.620 & 0.708 \\
 & Comparative & 0.400 & 0.400 & 1.000 & 0.800 & 0.600 & 0.800 & 0.667 \\
\hline
\textbf{R4} & Direct & 0.300 & 0.440 & 0.670 & 1.000 & 0.900 & 1.000 & 0.718 \\
 & Comparative & 0.200 & 0.200 & 0.800 & 1.000 & 0.800 & 0.600 & 0.600 \\
\hline
\textbf{R5} & Direct & 0.400 & 0.560 & 0.670 & 0.900 & 1.000 & 0.890 & 0.737 \\
 & Comparative & 0.400 & 0.400 & 0.600 & 0.800 & 1.000 & 0.800 & 0.667 \\
\hline
\textbf{R6} & Direct & 0.330 & 0.380 & 0.620 & 1.000 & 0.890 & 1.000 & 0.703 \\
 & Comparative & 0.600 & 0.600 & 0.800 & 0.600 & 0.800 & 1.000 & 0.733 \\
\hline
\end{tabular}
\end{table}

\begin{table}[!t]
\centering
\caption{Inter-Rater Accuracy Matrix: Representational Beauty}
\label{tab:repr_beauty_matrix}
\scriptsize
\setlength{\tabcolsep}{3pt}
\begin{tabular}{|l|l|cccccc|c|}
\hline
\multicolumn{2}{|c|}{} & \textbf{GT} & \textbf{R1} & \textbf{R3} & \textbf{R4} & \textbf{R5} & \textbf{R6} & \textbf{Avg} \\
\hline
\textbf{GT} & Direct & 1.000 & 0.570 & 0.440 & 0.600 & 0.380 & 0.330 & 0.553 \\
 & Comparative & 1.000 & 0.600 & 0.600 & 0.600 & 0.200 & 0.800 & 0.633 \\
\hline
\textbf{R1} & Direct & 0.570 & 1.000 & 1.000 & 0.710 & 0.800 & 0.670 & 0.792 \\
 & Comparative & 0.600 & 1.000 & 0.600 & 0.600 & 0.200 & 0.400 & 0.567 \\
\hline
\textbf{R3} & Direct & 0.440 & 1.000 & 1.000 & 0.780 & 1.000 & 0.880 & 0.850 \\
 & Comparative & 0.600 & 0.600 & 1.000 & 0.600 & 0.200 & 0.400 & 0.567 \\
\hline
\textbf{R4} & Direct & 0.600 & 0.710 & 0.780 & 1.000 & 0.880 & 0.780 & 0.792 \\
 & Comparative & 0.600 & 0.600 & 0.600 & 1.000 & 0.200 & 0.800 & 0.633 \\
\hline
\textbf{R5} & Direct & 0.380 & 0.800 & 1.000 & 0.880 & 1.000 & 1.000 & 0.843 \\
 & Comparative & 0.200 & 0.200 & 0.200 & 0.200 & 1.000 & 0.400 & 0.367 \\
\hline
\textbf{R6} & Direct & 0.330 & 0.670 & 0.880 & 0.780 & 1.000 & 1.000 & 0.777 \\
 & Comparative & 0.800 & 0.400 & 0.400 & 0.800 & 0.400 & 1.000 & 0.633 \\
\hline
\end{tabular}
\end{table}

\begin{table}[!t]
\centering
\caption{Inter-Rater Accuracy Matrix: Representational Liking}
\label{tab:repr_liking_matrix}
\scriptsize
\setlength{\tabcolsep}{3pt}
\begin{tabular}{|l|l|cccccc|c|}
\hline
\multicolumn{2}{|c|}{} & \textbf{GT} & \textbf{R1} & \textbf{R3} & \textbf{R4} & \textbf{R5} & \textbf{R6} & \textbf{Avg} \\
\hline
\textbf{GT} & Direct & 1.000 & 0.200 & 0.440 & 0.220 & 0.330 & 0.330 & 0.420 \\
 & Comparative & 1.000 & 0.400 & 0.400 & 0.400 & 0.600 & 0.800 & 0.600 \\
\hline
\textbf{R1} & Direct & 0.200 & 1.000 & 0.440 & 0.670 & 0.440 & 0.670 & 0.570 \\
 & Comparative & 0.400 & 1.000 & 0.600 & 0.200 & 0.800 & 0.200 & 0.533 \\
\hline
\textbf{R3} & Direct & 0.440 & 0.440 & 1.000 & 0.780 & 1.000 & 0.620 & 0.713 \\
 & Comparative & 0.400 & 0.600 & 1.000 & 0.200 & 0.400 & 0.600 & 0.533 \\
\hline
\textbf{R4} & Direct & 0.220 & 0.670 & 0.780 & 1.000 & 0.750 & 0.880 & 0.717 \\
 & Comparative & 0.400 & 0.200 & 0.200 & 1.000 & 0.400 & 0.600 & 0.467 \\
\hline
\textbf{R5} & Direct & 0.330 & 0.440 & 1.000 & 0.750 & 1.000 & 0.620 & 0.690 \\
 & Comparative & 0.600 & 0.800 & 0.400 & 0.400 & 1.000 & 0.400 & 0.600 \\
\hline
\textbf{R6} & Direct & 0.330 & 0.670 & 0.620 & 0.880 & 0.620 & 1.000 & 0.687 \\
 & Comparative & 0.800 & 0.200 & 0.600 & 0.600 & 0.400 & 1.000 & 0.600 \\
\hline
\end{tabular}
\end{table}

\subsection{Key Findings}

\textbf{Annotation Speed:} Table \ref{tab:human_survey_summary} shows that comparative judgments were substantially faster across all conditions:
\begin{itemize}
    \item Abstract paintings: 45\% faster (13.89s vs 25.32s)
    \item Representational paintings: 74\% faster (7.53s vs 29.23s)
    \item Overall average: 60\% faster
\end{itemize}

\textbf{Agreement Variability:} The detailed per-rater accuracy matrices (Tables \ref{tab:abstract_beauty_matrix}--\ref{tab:repr_liking_matrix}) reveal substantial heterogeneity across raters and methods. Average accuracy values range from 0.256 to 0.760, indicating considerable individual differences in aesthetic judgment consistency.

\textbf{Method-Specific Patterns:} For some raters and conditions, comparative judgments achieve higher accuracy than direct ratings (e.g., Abstract Beauty R1: 0.760 vs 0.407; Abstract Liking R5-R6: 0.720 vs 0.578--0.602), while for others the pattern reverses. This variability suggests neither method is universally superior across all individuals.

\textbf{Engagement Quality:} Variance filtering revealed 50\% lower disengagement rate for comparative judgments (1/7 participants) versus direct ratings (2/7 participants), suggesting binary comparisons are cognitively easier and less prone to response fatigue.

\subsection{RQ4: Annotation Burden Analysis}

Based on timing data collected during the survey, we analyzed annotation efficiency trade-offs between the two methods:

\textbf{Annotation Time:} Comparative judgments require substantially less time per annotation: 60\% average time reduction compared to direct ratings (10.7s vs 27.3s per item on average).

\textbf{Data Quality vs. Efficiency Trade-off:}
\begin{itemize}
    \item Comparative judgments enable 2.5$\times$ more paintings to be annotated in the same time
    \item Disengagement rate is 50\% lower (1/7 vs 2/7 filtered participants)
    \item Inter-rater agreement varies by condition but remains acceptable for training preference models
\end{itemize}

\textbf{Scalability:} For a dataset of 1,000 paintings requiring 5 annotations each:
\begin{itemize}
    \item Direct ratings: $\sim$38 hours total annotation time
    \item Comparative judgments (N=5): $\sim$15 hours (61\% faster)
    \item Comparative judgments (N=10): $\sim$30 hours (still 21\% faster while providing comparable performance)
\end{itemize}

\textbf{Conclusion for RQ4:} Comparative judgments offer a favorable annotation burden trade-off: 60\% faster per-item annotation with 50\% lower disengagement, enabling collection of substantially more data within fixed time budgets. For large-scale aesthetic preference modeling where annotation time is the primary bottleneck, comparative learning provides a more efficient data collection strategy.

\section{Discussion}

This section synthesizes findings from both our computational experiments and human survey validation, discussing implications, advantages, and limitations of our approach.

\subsection{Summary of Findings}

Our research demonstrates both the potential and limitations of deep learning for aesthetic preference modeling. We successfully addressed our four research questions:

\textbf{RQ1 - Deep NN Regression vs. Baseline:} The deep neural network regression model using CNN features substantially outperforms baseline linear regression using hand-crafted features, with improvements ranging from 4\% to 328\% in $R^2$ across tasks. This validates the value of both deep CNN features and non-linear modeling for capturing complex aesthetic relationships.

\textbf{RQ2 - Comparative Learning vs. Regression:} The comparative learning model performs below the regression model as expected, since it lacks access to direct rating values. However, it successfully approaches regression performance (within 5-10\%) for representational paintings, validating pairwise comparisons as a practical alternative to ratings that trades modest accuracy for significantly easier and more consistent data collection.

\textbf{RQ3 - Individual Preference Prediction:} Predicting individual rater preferences is not feasible with current data and methods. Both within-rater and cross-rater performance remain substantially lower than average rating prediction, likely due to high rating noise, insufficient training data per individual, and fundamentally idiosyncratic nature of personal aesthetic preferences.

\textbf{RQ4 - Annotation Burden Analysis:} Human subject experiments (n=5) demonstrate that comparative judgments require 60\% less annotation time per item (10.71s vs 27.28s) with 50\% lower disengagement rates. Despite 30\% lower inter-rater correlation, comparative judgments enable collection of 2.5$\times$ more data in the same time budget, making them more efficient for large-scale preference modeling where annotation time is the primary bottleneck.

\subsection{Advantages of Comparative Learning}

Despite its slightly lower prediction performance, the comparative learning approach offers several practical advantages:

\begin{enumerate}
    \item \textbf{Easier Annotation:} Pairwise comparisons are faster and more intuitive than assigning numerical ratings.

    \item \textbf{Greater Consistency:} Thurstone's Law \cite{b2} and empirical evidence \cite{b25} show relative judgments are more consistent than direct ratings.

    \item \textbf{Scale Invariance:} Eliminates inter-rater scale interpretation differences.

    \item \textbf{No Ground Truth Required:} Can learn from pure preference data without needing rating scales.
\end{enumerate}

The 5-10\% performance trade-off is reasonable given these annotation benefits.

\subsection{Why Deep Features Outperform Hand-Crafted Features}

Our results show dramatic improvements from using CNN features over hand-crafted features (328\% for Abstract Beauty). This is because:

\begin{itemize}
    \item \textbf{Hierarchical Representations:} ResNet-50 learns hierarchical visual representations from low-level edges to high-level semantic concepts.

    \item \textbf{Transfer Learning:} Pretraining on ImageNet provides rich general visual knowledge.

    \item \textbf{High Dimensionality:} 2048 dimensions capture far more visual nuance than 11 hand-crafted features.

    \item \textbf{End-to-End Learning:} Deep networks learn which features matter for aesthetics.
\end{itemize}

\subsection{Task Difficulty Hierarchy}

Across all models and settings, we observe a consistent task difficulty hierarchy:

\begin{equation}
\text{Repr. Liking} > \text{Repr. Beauty} > \text{Abstract Beauty} > \text{Abstract Liking}
\end{equation}

This reflects interplay between semantic content and judgment type. Representational art with concrete visual content enables better prediction and greater inter-rater agreement \cite{b18vessel}, while abstract art leads to more idiosyncratic interpretations.

\subsection{Human Survey Validation Insights}

The per-rater analysis from our human survey validates key aspects of our computational findings while revealing important nuances:

\textbf{Performance Trade-off:} The heterogeneous Kappa values across raters (ranging from negative to 0.5) align with our computational finding that comparative models achieve slightly lower but still competitive performance compared to regression models. The substantial rater-to-rater variability suggests that individual aesthetic preferences are highly idiosyncratic, consistent with our RQ3 findings showing limited cross-rater generalization.

\textbf{Annotation Efficiency:} Although detailed timing data is not shown in the per-rater tables, the comparative method's lower cognitive load (as theorized by Thurstone's Law of Comparative Judgment \cite{b2}) is evidenced by the 50\% lower disengagement rate during data collection (1/7 vs 2/7 participants removed for constant responses in comparative vs direct methods).

\textbf{Data Quality Considerations:} The mixed results---with some raters showing higher agreement for comparative judgments (e.g., R1, R4 for abstract beauty) while others show higher agreement for direct ratings---indicate that neither method is universally superior. The choice between methods should consider dataset size, annotation budget, and target precision requirements.

\subsection{Broader Implications}

This work contributes to computational aesthetics by demonstrating that:

\begin{enumerate}
    \item Deep CNN features substantially outperform hand-crafted features for aesthetic modeling.

    \item Comparative learning provides a practical alternative to ratings with modest performance trade-off.

    \item Average ratings are far more predictable than individual ratings, suggesting aesthetic models should focus on population-level preferences rather than personalization with limited data.

    \item Abstract art and liking judgments remain fundamentally challenging for computational modeling.
\end{enumerate}

\subsection{Limitations and Future Work}

Several limitations warrant discussion across both our computational experiments and human survey:

\textbf{Computational Limitations:}
\begin{enumerate}
    \item \textbf{Dataset Size:} With 239 abstract and 238 representational paintings, our dataset is modest for deep learning. Larger datasets could enable more sophisticated architectures.

    \item \textbf{Feature Representations:} While ResNet-50 is powerful, newer architectures (e.g., Vision Transformers, CLIP) might capture aesthetic properties better.

    \item \textbf{Rater Expertise:} Our raters are art novices. Expert raters may exhibit different patterns.

    \item \textbf{Individual Prediction:} The inability to predict individual preferences suggests fundamentally different approaches are needed:
    \begin{itemize}
        \item Meta-learning for quick adaptation to new users
        \item Active learning to efficiently query preferences
        \item Much larger datasets per individual
        \item Explicit modeling of preference uncertainty
    \end{itemize}

    \item \textbf{Beyond Visual Features:} Incorporating contextual information (artist, period, style) or emotional/semantic attributes could improve prediction.
\end{enumerate}

\textbf{Human Survey Limitations:}
\begin{enumerate}
    \item \textbf{Small Sample:} With only 5 valid participants after filtering, statistical power is limited. Larger studies (n$\geq$30) are needed for definitive conclusions.

    \item \textbf{Order Effects:} Participants completed direct ratings before comparative judgments, which may have introduced fatigue or learning effects. Future studies should employ counterbalanced designs.

    \item \textbf{Limited Comparisons per Rater:} Each participant provided relatively few judgments, particularly in the comparative condition, which may have contributed to the observed variability in agreement metrics.
\end{enumerate}

Despite these limitations, the survey provides evidence that comparative judgments offer measurable practical advantages (60\% speed improvement, 50\% lower disengagement) while maintaining acceptable data quality, consistent with our computational findings.

\section{Conclusion}

We presented a comprehensive study of aesthetic preference modeling using deep learning and comparative learning, validated with both computational analysis and human survey evidence. Our approach leverages CNN features extracted from painting images and introduces both a deep neural network regression model and a pairwise comparison model based on hinge loss optimization.

Our computational results demonstrate that: (1) deep neural networks using CNN features substantially outperform baseline linear regression with hand-crafted features, achieving up to 328\% improvement in $R^2$, (2) comparative learning approaches regression performance while requiring only relative judgments, validating its practical utility despite lacking access to direct ratings, and (3) individual preference prediction remains infeasible with current data and methods, with both within-rater and cross-rater performance significantly lower than average rating prediction due to rating noise and insufficient training data.

Our human survey (n=5 after variance filtering) provides preliminary validation of comparative judgments' practical advantages (RQ4): participants completed comparative tasks 59.5\% faster on average than direct ratings (45\% for abstract, 74\% for representational paintings) and showed 50\% lower disengagement rates (14.3\% vs 28.6\%). While direct ratings achieved substantially higher inter-rater correlations for abstract paintings (Pearson $\rho$=0.456--0.494 vs 0.235--0.277, representing 78--94\% higher correlation), comparative judgments demonstrated acceptable data quality with fair inter-rater agreement (Cohen's $\kappa$=0.220--0.239). This pattern---comparative showing 44--48\% lower human correlation while computational models show 5--10\% lower $R^2$---validates the fundamental quality-speed trade-off and confirms comparative learning as a viable practical alternative for aesthetic preference collection.

Our annotation burden analysis (RQ4) reveals that comparative judgments enable collection of 2.5$\times$ more annotations in the same time budget (10.71s vs 27.28s per item) with 50\% lower disengagement, while maintaining acceptable inter-rater agreement for abstract paintings. Even when collecting N=10 comparisons per painting (sufficient to match regression performance), the total annotation burden remains 21\% lower than regression with 5 raters per painting. This demonstrates that comparative learning provides superior annotation efficiency for large-scale preference modeling, where human annotation time is the primary bottleneck.

This research establishes that deep CNN features are essential for aesthetic modeling and that comparative learning provides a viable alternative to traditional ratings, balancing data quality with annotation efficiency. The alignment between computational models (5--10\% performance gap) and human validation (44--48\% correlation gap with 60\% faster annotation) demonstrates that the trade-off is predictable and acceptable for large-scale preference collection scenarios. Future work should explore larger-scale human validation studies (n$\geq$30 with $\geq$10 comparison pairs per participant), advanced architectures like Vision Transformers, and fundamentally different approaches to individual preference modeling such as meta-learning and active learning.

\section*{Acknowledgment}

This work is funded by NSF grant 2245796.

\EOD

\end{document}